\documentclass[10pt,twocolumn,letterpaper]{article}

\usepackage{algorithm}
\usepackage{algorithmic}
\usepackage{amsmath}
\usepackage{amssymb}
\usepackage{multirow}
\usepackage{svg}

\usepackage{cvpr}

\definecolor{linkblue}{rgb}{0.21,0.49,0.74}
\usepackage[pagebackref,breaklinks,colorlinks,allcolors=linkblue]{hyperref}

\def\paperID{}
\def\confName{}
\def\confYear{}

\title{TokenFlow: Latent Token Rectified Flow Matching \\ for 3D Point Cloud Generation}

\author{
Linhao Wang$^{1}$, Dongjin Huang$^{1, \dag}$\\
$^{1}$Shanghai Film Academy, Shanghai University \\
$^{\dag}$ Corresponding Author
}

\begin{document}
\maketitle
\begin{abstract}
3D point clouds generation requires capturing both part-level layout and surface detail while preserving permutation equivariance. 
Existing latent generative models often encode an entire shape into a single vector, which limits their ability to retain part-level structure. 
We propose TokenFlow, a hierarchical flow matching framework that represents each shape as a set of latent tokens and couples a rectified flow over these tokens with a conditional flow over points. 
A bottleneck of $M \ll N$ learnable tokens aggregates global context through cross-attention at $O(NM)$ cost. 
On ShapeNet, TokenFlow achieves the best MMD and coverage across airplane, car, and chair, while obtaining 1-NNA comparable to the strongest baselines. 
On ModelNet, its learned latent set reaches the best reported linear-evaluation accuracy on ModelNet10. 
These results show that representing a shape as a set of latent tokens, aggregated at $O(NM)$ cost, provides an effective representation for 3D point cloud generation.
\end{abstract}    
\section{Introduction}
\label{sec:intro}

A 3D shape is more than a collection of points. Although a point cloud is represented as an unordered set of coordinates, its identity is largely determined by how parts are organized and how their surfaces are formed. A chair, for example, requires the seat, back, and legs to occupy a coherent spatial configuration, while each individual leg must still exhibit a well-defined local surface. These two requirements are fundamentally different: \emph{global structure} determines where parts should be and how they relate to one another, whereas \emph{local geometry} determines how those parts are realized at the point level. A useful generative model must capture both while remaining permutation-equivariant.

Point cloud generation has progressed substantially along several directions. Early GAN- and VAE-based approaches \cite{achlioptas2018learning,li2018point,gadelha2018multiresolution,kim2021setvae,wang2024dpr} directly generate point sets and optimize set distances such as Chamfer Distance (CD) and Earth Mover's Distance (EMD) \cite{fan2017point}. Continuous normalizing flows (CNFs), including PointFlow \cite{yang2019pointflow}, SoftFlow \cite{kim2020softflow}, and DPF-Net \cite{klokov2020discrete}, instead model point clouds through continuous transformations of distributions. More recently, diffusion models \cite{ho2020denoising,luo2021diffusion,zhou20213d,vahdat2022lion} have improved generation quality, while latent representations \cite{vahdat2022lion,zhang20233dshape2vecset,mo2023dit,koo2023salad} and flow matching \cite{lipman2022flow,liu2022flow} have emerged as effective ways to improve the scalability and efficiency of generative modeling. Despite these advances, most existing models still treat the generation process as a single prediction problem: one representation and one generative field are responsible for both global organization and local surface formation.

This design creates a less obvious bottleneck. Consider a rectified transport path

$$
Y_t=(1-t)Y_0+tX.
$$

At an early time $t$, the current state contains little information about the target shape; later in the trajectory, the target geometry becomes increasingly identifiable. Thus, the useful signal available for predicting the target is not uniform throughout the transport process. Global structure and local geometry are also not equally informative at the same time, and the two types of information cannot be expected to share a single representation.

The representation used by latent generative models makes this problem more pronounced. Recent approaches such as LION \cite{vahdat2022lion} and 3DShape2VecSet \cite{zhang20233dshape2vecset} avoid operating directly on all points by first mapping a shape into a latent representation. However, compressing an entire shape into a single global vector makes part-level structure difficult to preserve explicitly. A natural alternative is to retain a set of latent tokens, so that different tokens can encode different aspects of the shape while remaining permutation-invariant as a whole. The challenge is then to connect this latent representation back to $N$ points without resorting to full point-to-point attention, whose $O(N^2)$ complexity becomes costly as the point resolution grows.

These observations suggest a simple question: \emph{where should the boundary between global organization and local geometry lie?} We answer this question with \textbf{TokenFlow}, a hierarchical flow matching framework that places the two roles at different levels while allowing them to interact throughout the transport process. TokenFlow represents a shape as a set of $M\ll N$ latent tokens rather than a single global code. These tokens provide a compact global state from which part-level organization can be modeled, while cross-attention connects the global representation back to individual points at $O(NM)$ cost.

Collectively, these components yield a generative model that disentangles shape semantics from surface geometry across different levels, while maintaining their coupling via the flow. We additionally conduct controlled experiments to isolate the effect of the representation.

Our contributions are summarized as follows:

\begin{itemize}

\item We introduce \textbf{TokenFlow}, a latent-set rectified flow framework for 3D point cloud generation that separates global shape organization from point-level detail in its representation.

\item We propose \textbf{context bottleneck aggregation}. Placing it after point-wise feature extraction means cross-attention learns only a grouping rather than a metric, and using asymmetric query/key versus value widths keeps the $O(NM)$ attention map low-dimensional while the carried information stays high-dimensional.

\item We evaluate TokenFlow on ShapeNet and ModelNet datasets, demonstrating superior performance compared with previous state-of-the-art methods.

\end{itemize}
\section{Related Work}
\subsection{Point Cloud Generation}
\label{sec:rw-gen}

Early point cloud generators directly model an $N\times3$ point matrix with GANs \cite{achlioptas2018learning,li2018point,wang2024dpr}, VAEs \cite{gadelha2018multiresolution,kim2021setvae}, and adversarial autoencoders \cite{zamorski2018adversarial,makhzani2015adversarial}. These methods typically use a fixed number of output points and rely on set distances such as Chamfer Distance and Earth Mover's Distance \cite{fan2017point}. FoldingNet \cite{yang2018foldingnet} and AtlasNet \cite{groueix2018papier} instead generate surfaces by deforming 2D patches. PointFlow \cite{yang2019pointflow} introduced a continuous normalizing flow formulation that models both the shape distribution and the conditional point distribution, allowing density estimation and variable point counts. SoftFlow \cite{kim2020softflow}, DPF-Net \cite{klokov2020discrete}, and ShapeGF \cite{cai2020learning} further explored continuous distributional formulations.

Diffusion models subsequently became a dominant paradigm for point cloud generation \cite{luo2021diffusion,zhou20213d,ren2024tiger} and latent 3D generation \cite{vahdat2022lion,mo2023dit,koo2023salad,Liu2023MeshDiffusion}. Recent work has extended diffusion to unified point cloud processing \cite{du2025superpc}, joint geometry-semantic generation \cite{stone2025guided}, and state-space architectures \cite{zhang2025enhancing}. Autoregressive models have also shown competitive performance through coarse-to-fine next-scale prediction \cite{meng2026pointnsp}. Beyond generation, diffusion features have been distilled for point cloud representation learning \cite{chen2025harnessing}. Our work instead separates the two at the level of representation: a set of latent tokens carries global shape organization, while a conditional flow resolves local geometry.
\begin{figure*}[!tb]
    \centering
    \includegraphics[width=\textwidth]{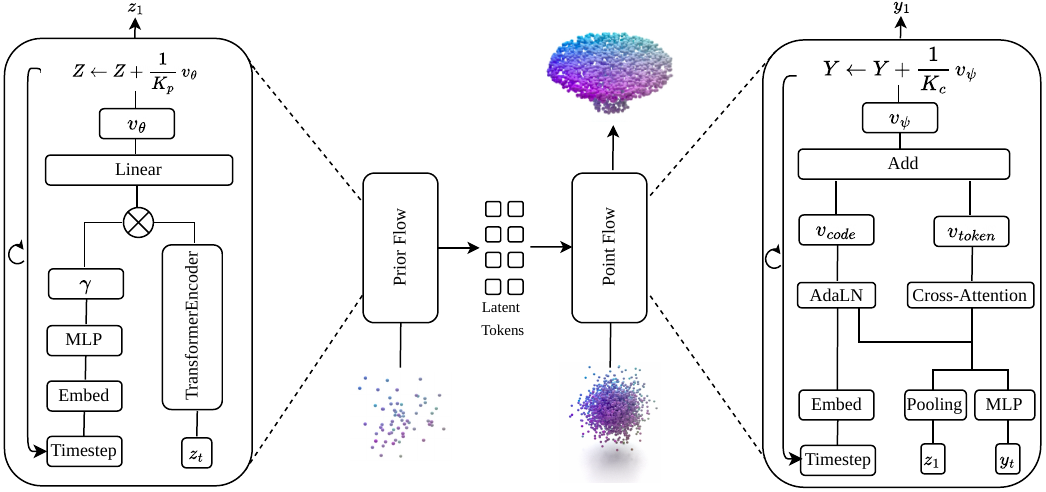}
    \caption{TokenFlow pipeline. The prior flow transports Gaussian noise to a set of
    latent tokens over a Transformer encoder with several layers whose output is rescaled by a
    time-conditioned $\gamma$. The conditional point flow then decodes those tokens
    into a point cloud. A shared per-point MLP embeds each point and feeds two
    separately projected branches that are added, a code path in which the pooled
    token set and the time embedding give the scale and shift of an adaptive layer
    normalization over the point features, and a token path that cross-attends from
    the points to the latent tokens.}
    \label{fig:main}
\end{figure*}
\subsection{Continuous-Time Generative Modeling}
\label{sec:rw-paradigm}

Continuous-time generative models provide a principled framework for transporting a simple reference distribution to the data distribution. Diffusion models \cite{ho2020denoising} perform this transformation through a predefined noise schedule, but typically require many denoising steps at inference. Flow matching \cite{lipman2022flow} directly learns a velocity field along a prescribed probability path, while rectified flow \cite{liu2022flow} uses straight interpolation paths that enable efficient Euler sampling. Optimal-transport couplings \cite{tong2023improving} further improve path straightness, and consistency models \cite{song2023consistency} reduce the number of required sampling steps.

These formulations have recently been applied to 3D generation. TripoSG \cite{li2025triposg} uses a large-scale rectified-flow transformer for image-to-3D synthesis, while ShapeR \cite{siddiqui2026shaper} conditions rectified flow on sparse SLAM points, posed images, and text. Related approaches target dynamic meshes \cite{wu2026r} and 3D Gaussian objects \cite{wang2026fusion}. Hierarchical Flow Matching \cite{wang2026hierarchical} factorizes point cloud generation into a latent flow and a conditional point flow. Other recent studies have revisited transport design and evaluation for point cloud generation \cite{hui2025not,bastico2026rethinking,vercheval2024pcgen}.

TokenFlow is closest to hierarchical flow-based formulations, but differs in how the shape is represented and how the conditional field is conditioned. It aggregates context through a bottleneck of $M\ll N$ latent tokens, which condition a point-level rectified flow at $O(NM)$ cost.

\subsection{Latent Shape Representations}
\label{sec:rw-repr}

The choice of representation strongly affects how well a generative model preserves 3D structure. Global latent vectors \cite{yang2019pointflow,luo2021diffusion} provide compact shape representations but compress part-level information into a single code. Voxel- and grid-based representations preserve spatial locality but incur rapidly increasing memory costs with resolution.

Latent sets provide an alternative by representing a shape with multiple learned tokens. LION \cite{vahdat2022lion} models a latent point cloud obtained from a hierarchical VAE, while 3DShape2VecSet \cite{zhang20233dshape2vecset} represents shapes as sets of latent vectors and applies a diffusion transformer for unconditional, category-, text-, and image-conditioned generation. Part-aware representations have also been explored in SPAGHETTI \cite{hertz2022spaghetti} and SALAD \cite{koo2023salad}. Autoregressive approaches instead discretize latent representations and generate them sequentially \cite{sun2020pointgrow,cheng2022autoregressive,chen2023pointgpt}. At the scene level, class-partitioned VQ-VAEs have recently been combined with latent flow matching for multi-object point cloud generation \cite{de2026class}.

TokenFlow also adopts a latent set, but uses it as explicit global context for a conditional point flow rather than as the input to a single monolithic denoiser. This separation allows coarse layout and point-level detail to be modeled at different levels of the transport process.

\subsection{Context Aggregation and Scalability on Point Sets}

\label{sec:rw-ctx}
Global context is essential for point cloud modeling, but direct point-to-point interaction scales poorly with the number of points. PointNet \cite{qi2017pointnet} uses symmetric pooling to obtain permutation invariance with negligible interaction cost, but the resulting global representation is not explicitly adaptive to individual points. Transformer-based architectures, including PCT \cite{guo2021pct} and Point Transformer \cite{zhao2021point,wu2022point}, model point-to-point relations through attention, with full attention requiring $O(N^2)$ time and memory. Local attention \cite{zhao2021point} reduces the interaction range but remains dependent on the point resolution.

A different strategy is to introduce a smaller set of latent tokens as an information bottleneck. Perceiver \cite{jaegle2021perceiver} cross-attends from a small array of learned latents to a large input and then processes the latents with self-attention, avoiding any $N \times N$ map. Hierarchical masked autoencoders \cite{zhang2022point} and 3DShape2VecSet \cite{zhang20233dshape2vecset} adopt the same recipe for 3D shapes. 
TokenFlow follows this bottleneck design and aggregates $N$ point features into $M \ll N$ context tokens through cross-attention, reducing the interaction cost to $O(NM)$. 
Our contribution here is not the bottleneck itself, which is established, but its placement and parametrisation: we place it after point-wise feature extraction, where cross-attention has to learn only a grouping rather than a metric, and we make the query/key and value widths asymmetric so that the $O(NM)$ attention map is constructed in a low-dimensional space. Neither choice is analysed in the works above; we quantify the resulting cost against global, local and linear attention in Tab.~\ref{tab:complexity}.

\section{Preliminaries}
\begin{figure*}[!tb]
    \centering
\includegraphics[width=\textwidth]
    {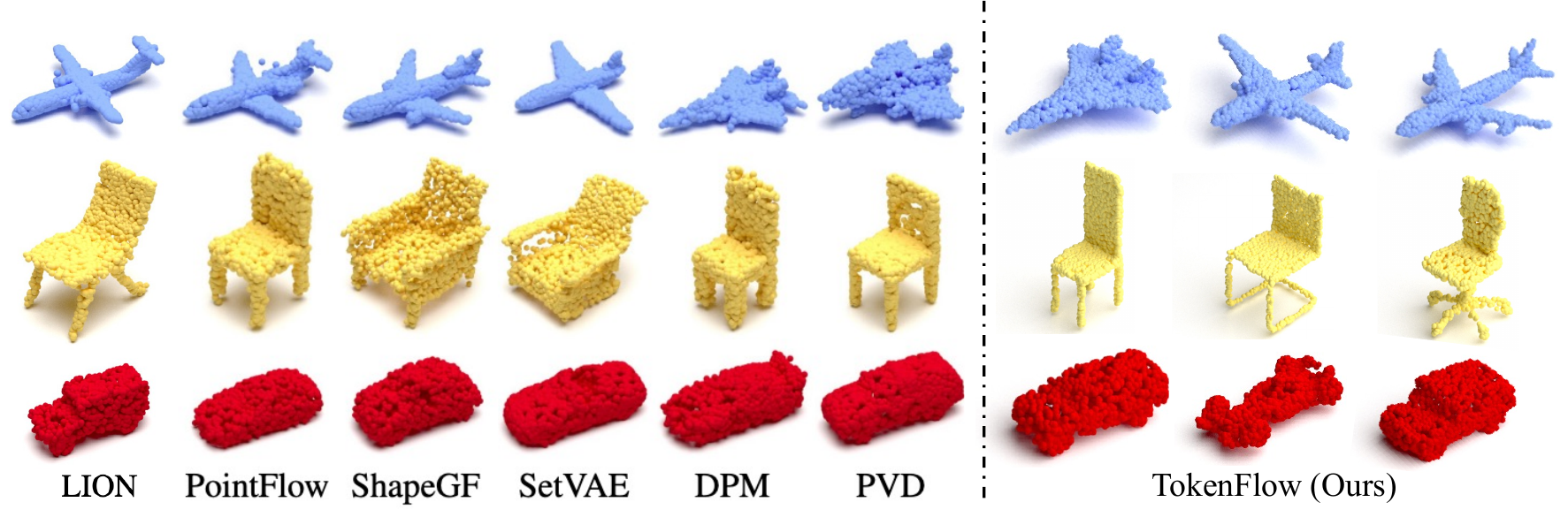}
    \caption{Qualitative comparisons against previous state-of-the-art methods.}
    \label{fig:gen}
\end{figure*}
\subsection{Continuous Normalizing Flows}
A continuous normalizing flow (CNF) \cite{chen2018neural,grathwohl2018ffjord} transports a noise sample $z(t_0)$ to the data space by integrating an ODE
\begin{align}
\frac{d z(t)}{d t} = f(z(t), t),
\end{align}
where $f$ is a learned vector field. Exact densities are available through the change-of-variables identity, but training pays for them three times over: the Jacobian trace must be estimated at every step, $f$ is confined to tractable forms such as affine couplings, and gradients must flow through the ODE solver itself. These costs are why point cloud generators built on CNFs \cite{yang2019pointflow,klokov2020discrete,kim2020softflow} have largely given way to simulation-free training.

\subsection{Rectified Flow Matching}
\label{sec:cfm}
Flow matching \cite{lipman2022flow} takes the opposite bargain: give up exact densities, and in exchange train by plain regression. Given data $a_1$ and noise $a_0 \sim \mathcal{N}(0, I)$, define the straight interpolation
\begin{align}
a_t = (1-t)\,a_0 + t\,a_1, \qquad t \in [0,1],
\label{eq:path}
\end{align}
and regress a network $v_\vartheta$ on the constant velocity along the segment,
\begin{align}
\mathcal{L}_{\mathrm{FM}} = \mathbb{E}\Big\| v_\vartheta(a_t, t) - (a_1 - a_0) \Big\|^2 .
\label{eq:fm}
\end{align}
Sampling integrates $d a_t / d t = v_\vartheta(a_t, t)$ from $t=0$ to $1$. Since each path has constant velocity, trajectories stay close to straight lines, and a handful of Euler steps is enough \cite{liu2022flow,tong2023improving}. Nothing needs to be invertible, so $v_\vartheta$ can be any network; in our case it is a permutation-equivariant one. Conditioning is free: a context $c$ simply enters as another argument of $v_\vartheta$, and Eq.~\eqref{eq:fm} becomes regression on $a_1 - a_0$ given $(a_t, t, c)$.

\subsection{Two-Level Generation}
A point cloud is itself a set of samples, which motivates the hierarchy of PointFlow \cite{yang2019pointflow}: a shape code $Z$ is drawn from a prior $P_\eta(Z)$, and the $N$ points are drawn conditionally, $X \sim P_\alpha(X|Z)$. Training maximizes the ELBO
\begin{align}
\log P_\alpha(X) \geq \mathbb{E}_{Q_\varphi(Z|X)}\!\big[\log P_\alpha(X|Z)\big] - \notag \\
D_{KL}\!\big(Q_\varphi(Z|X) \,\parallel\, P_\eta(Z)\big).
\label{eq:ELBO}
\end{align}
We keep this hierarchy and change both of its ingredients: $Z$ becomes a \emph{set} of $M$ tokens rather than one vector, and both levels are learned by the regression of Eq.~\eqref{eq:fm} instead of by CNF likelihoods.

\section{Method}
\subsection{Overview}
A point cloud is denoted as $X = \{x_i\}_{i=1}^{N}$, where $x_i \in \mathbb{R}^{3}$, and a shape is represented by a compact set of tokens $Z = \{z_m\}_{m=1}^{M}$, where $z_m \in \mathbb{R}^{d}$ and $M \ll N$. TokenFlow jointly trains two components (Fig.~\ref{fig:main}): a latent-set autoencoder $Q_\varphi(Z|X)$, which maps the input point cloud to a compact token set and reconstructs the point cloud from it (Sec.~\ref{sec:repr}), and two rectified flow models, including a prior over the token set and a conditional flow that generates point coordinates conditioned on $Z$ (Secs.~\ref{sec:rfm_prior} and~\ref{sec:rfm}). The overall training objective is described in Sec.~\ref{sec:obj}.

\begin{figure*}[!tb]
    \centering
\includegraphics[width=\textwidth]
    {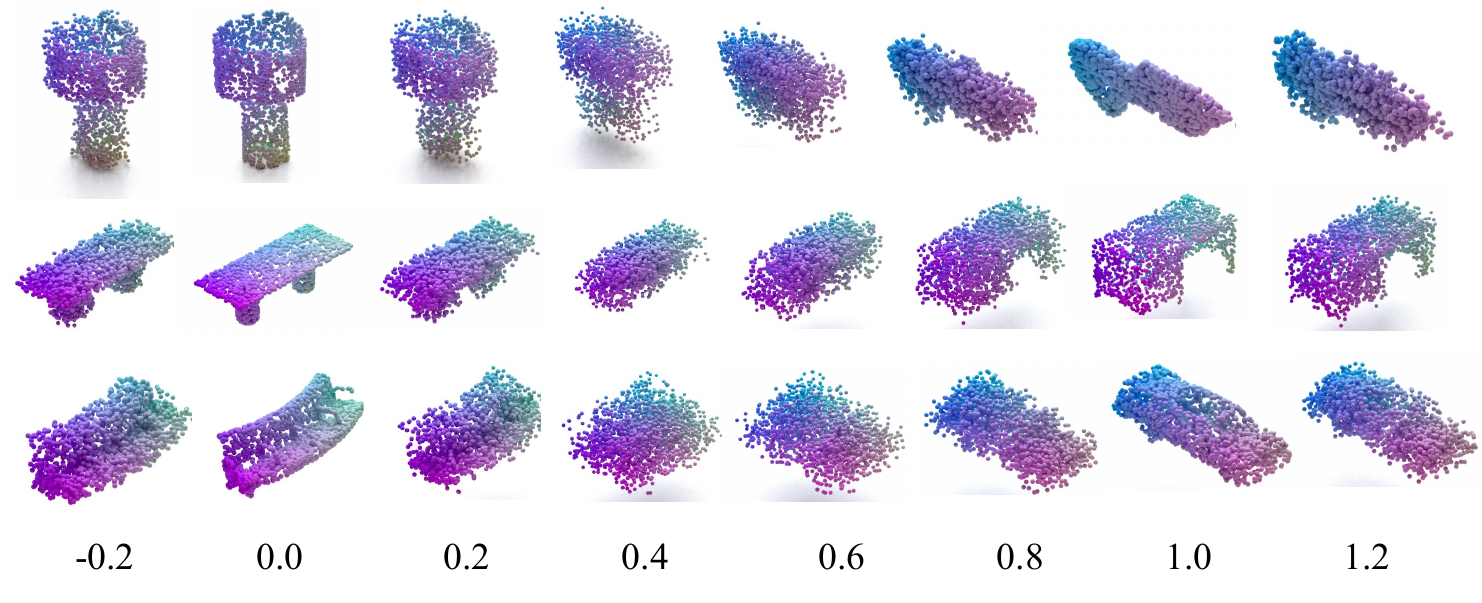}
    \caption{Interpolation and extrapolation in the latent space.}
    \label{fig:inter_extra}
\end{figure*}

\begin{table}[t]
\centering
\scriptsize
\begin{tabular}{lcc}
\toprule
Context module & Time & Memory \\
\midrule
Transformer \cite{vaswani2017attention} & $O(N^2 C)$ & $O(N^2)$ \\
PointTransformer \cite{zhao2021point} & $O(N k C)$ & $O(N k)$ \\
Performer \cite{choromanski2020rethinking} & $O(N C^2)$ & $O(N C)$ \\
CBA (ours) & $O(N M C)$ & $O(N M)$ \\
\bottomrule
\end{tabular}
\caption{Complexity of context aggregation over $N$ points with $C$ channels. CBA uses $M \ll N$ latent tokens, so both time and memory are linear in $N$.}
\label{tab:complexity}
\end{table}
\begin{figure}[htbp]
  \centering
  \includegraphics[width=0.4\columnwidth]{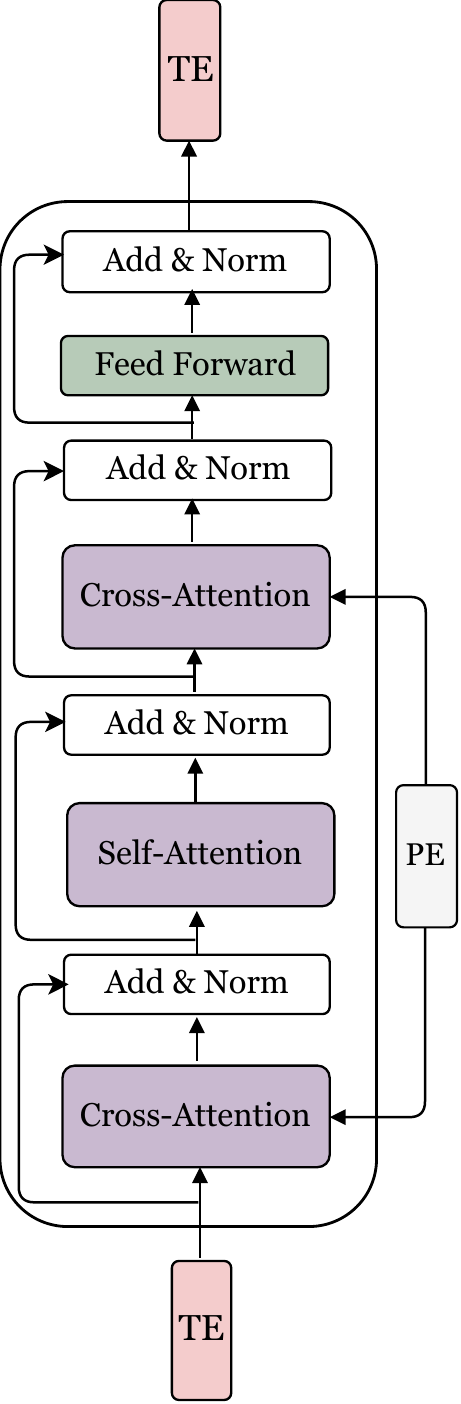}
  \caption{Illustration of CBA block. "TE" means token embeddings while "PE" represents point embeddings.}
  \label{fig:cba}
\end{figure}
\subsection{Latent-Set Shape Encoder}
\label{sec:repr}
\paragraph{\textbf{Point-wise geometry embedding.}}
Each point is independently embedded by a shared $1{\times}1$ convolution,
$X \in \mathbb{R}^{B \times N \times 3} \mapsto H^{(0)} \in \mathbb{R}^{B \times N \times C}$, with $C=128$. This operation is permutation equivariant with $O(NC)$ complexity but captures no inter-point interactions, which are introduced by the bottleneck below.
\paragraph{\textbf{Context Bottleneck Aggregation.}}
\label{sec:cba}
Transformer self-attention over all $N$ points costs $O(N^2)$ time and memory. We instead aggregate through $M$ learnable tokens $Q = \{q_m\}_{m=1}^{M}$ that cross-attend to the point features. Each block (Fig.~\ref{fig:cba}) computes
\begin{align}
Z = \mathrm{CBA}(H),
\label{eq:cba}
\end{align}
where $\mathrm{CBA}(\cdot)$ denotes a stack of $L_b$ bottleneck blocks. Inside each block, point features $H \in \mathbb{R}^{B \times N \times C}$ are first aggregated into $M$ latent tokens by a cross-attention layer,
\begin{equation}
\tilde{Z} = \mathrm{LN}\!\left( Q + \mathrm{softmax}\!\left( \frac{(Q W_Q)(H W_K)^{\top}}{\sqrt{C}} \right) (H W_V) \right),
\end{equation}
followed by a self-attention layer,
\begin{equation}
\hat{Z} = \mathrm{LN}\!\left( \tilde{Z} + \mathrm{SelfAttn}(\tilde{Z}) \right),
\end{equation}
then a second cross-attention in which the updated tokens re-read the point features for calibration,
\begin{equation}
Z' = \mathrm{LN}\!\left( \hat{Z} + \mathrm{softmax}\!\left( \frac{(\hat{Z} W_Q')(H W_K')^{\top}}{\sqrt{C}} \right) (H W_V') \right),
\end{equation}
and finally a feed-forward network,
\begin{equation}
Z'' = \mathrm{LN}\!\left( Z' + \mathrm{FFN}(Z') \right).
\end{equation}
The block output is $Z''$, and every sublayer applies layer normalization after adding its residual. Here $Q \in \mathbb{R}^{M \times C}$ are the learnable latent tokens and $\mathrm{LN}(\cdot)$ denotes layer normalization \cite{ba2016layer}.

Two design choices are important. First, we apply the bottleneck after the point-wise embedding rather than directly to the raw coordinates. Each point therefore already has a $C$-dimensional feature that captures local geometry, allowing the cross-attention to focus on grouping points instead of learning low-level geometric relations from scratch. Second, we use asymmetric projection dimensions. Following \cite{vaswani2017attention}, the query and key dimensions are set to $C/8$, while the value dimension remains $C$. This keeps the $O(NM)$ attention map low-dimensional while preserving the full feature dimension for information propagation.

Tab.~\ref{tab:complexity} summarizes the cost. At $N = 2048$ and $M = 64$ the pairwise interactions fall from $N^2 \approx 4.2 \times 10^{6}$ to $NM \approx 1.3 \times 10^{5}$, over an order of magnitude, and attention memory becomes linear in $N$.

\begin{table*}[t]
\centering
\scalebox{0.83}{
\begin{tabular}{l|ccccc|ccccc|ccccc}
\toprule
\multirow{3}{*}{Method}
& \multicolumn{5}{c}{Chair}
& \multicolumn{5}{c}{Airplane}
& \multicolumn{5}{c}{Car} \\
\cmidrule(lr){2-6} \cmidrule(lr){7-11} \cmidrule(lr){12-16}
& \multicolumn{2}{c}{MMD $\downarrow$}
& \multicolumn{2}{c}{COV(\%) $\uparrow$}
& JSD $\downarrow$
& \multicolumn{2}{c}{MMD $\downarrow$}
& \multicolumn{2}{c}{COV(\%) $\uparrow$}
& JSD $\downarrow$
& \multicolumn{2}{c}{MMD $\downarrow$}
& \multicolumn{2}{c}{COV(\%) $\uparrow$}
& JSD $\downarrow$ \\

& CD & EMD & CD & EMD & --
& CD & EMD & CD & EMD & --
& CD & EMD & CD & EMD & -- \\
\midrule

r-GAN
& 2.57 & 12.80 & 33.99 & 9.97 & 11.50
& 0.261 & 5.47 & 42.72 & 18.02 & 7.44
& 1.27 & 8.74 & 15.06 & 9.38 & 12.80 \\

l-GAN (CD)
& 2.46 & 8.91 & 41.39 & 25.68 & 4.59
& 0.239 & 4.27 & 43.21 & 21.23 & 4.62
& 1.55 & 6.25 & 38.64 & 18.47 & 4.43 \\

l-GAN (EMD)
& 2.61 & \underline{7.85} & 40.79 & 41.69 & 2.27
& 0.269 & 3.29 & 47.90 & 50.62 & \textbf{3.61}
& 1.48 & 5.43 & 39.20 & 39.77 & 2.21 \\

PVD
& 2.62 & -- & \underline{48.84} & \underline{50.60} & --
& 0.224 & -- & \underline{48.88} & \underline{52.09} & --
& 1.10 & -- & 41.19 & \underline{50.56} & -- \\
GET3D
& -- & -- & 43.36 & 42.77 & --
& -- & -- & -- & -- & --
& -- & -- & 15.04 & 18.38 & -- \\

DPF-Net
& 2.54 & -- & 44.71 & 48.79 & --
& 0.264 & -- & 46.17 & 48.89 & --
& 1.13 & -- & \underline{45.74} & 49.43 & -- \\

SoftFlow
& 2.53 & -- & 41.39 & 47.43 & --
& 0.231 & -- & 46.91 & 47.90 & --
& 1.19 & -- & 42.90 & 44.60 & -- \\

PointFlow 
& \underline{2.42} & 7.87 & 46.83 & 46.98 & \underline{1.74}
& \underline{0.217} & \underline{3.24} & 46.91 & 48.40 & 4.92
& \underline{0.91} & \underline{5.22} & 44.03 & 46.59 & \underline{0.87} \\

PC-GAN
& 2.75 & 8.20 & 36.50 & 38.98 & 3.90
& 0.287 & 3.57 & 36.46 & 40.94 & 4.63
& 1.12 & 5.83 & 23.56 & 30.29 & 5.85 \\

MeshDiffusion
& -- & -- & 46.00 & 46.71 & --
& -- & -- & 47.34 & 42.15 & --
& -- & -- & 34.07 & 25.85 & -- \\
\midrule
TokenFlow (ours)
& \textbf{2.35} & \textbf{7.63} & \textbf{51.28} & \textbf{51.37} & \textbf{1.51}
& \textbf{0.211} & \textbf{3.07} & \textbf{52.91} & \textbf{52.72} & \underline{4.52}
& \textbf{0.88} & \textbf{5.12} & \textbf{49.75} & \textbf{52.76} & \textbf{0.75} \\

\bottomrule
\end{tabular}
}
\caption{Generation results using MMD, COV and JSD ($\uparrow$/$\downarrow$ as marked). MMD-CD and MMD-EMD are scaled by $10^3$ and $10^2$; JSD by $10^2$. “--” marks unreported entries.}
\label{tab:other_metric_results}
\end{table*}

\begin{table}[t]
\centering
\scalebox{0.9}{
\begin{tabular}{lcc}
\toprule
Model & MN10 (\%) & MN40 (\%) \\
\midrule
SPH & 68.2 & 79.8 \\
LFD & 75.5 & 79.9 \\
T-L Network & 74.4 & --   \\
Vconv-DAE & 75.5 & 80.5 \\
PointGrow    & 85.7 & --   \\
3D-GAN       & 83.3 & 91.0 \\
PointFlow    & 86.8 & 93.7 \\
DPM          & 87.6 & \textbf{94.2} \\
DualGAN      & 87.6  & 94.1 \\
TGN          & \underline{87.7} & \textbf{94.2} \\
\midrule
TokenFlow (ours) & \textbf{87.9} & \underline{93.9} \\
\bottomrule
\end{tabular}
}
\caption{Classification results on ModelNet10 (MN10) and ModelNet40 (MN40) datasets. Missing entries are denoted as “–”.}
\label{tab:modelnet_cls}
\end{table}

\begin{table*}
    \centering
    \scalebox{1}{
    \begin{tabular}{lcccccc} 
        \toprule
        \multirow{2}{*}{Method} & \multicolumn{2}{c}{Airplane} & \multicolumn{2}{c}{Car} & \multicolumn{2}{c}{Chair} \\
        \cmidrule(lr){2-3} \cmidrule(lr){4-5} \cmidrule(lr){6-7} & CD & EMD & CD & EMD & CD & EMD \\
        \midrule
        r-GAN  & 98.40 & 96.79 & 94.46 & 99.01 & 83.69 & 99.70 \\
        l-GAN (CD)  & 87.30 & 93.95 & 66.49 & 88.78 & 68.58 & 83.84 \\
        l-GAN (EMD)  & 89.49 & 76.91 & 71.16 & 66.19 & 71.90 & 64.65 \\
        PC-GAN  & 94.35 & 92.32 & 92.19 & 90.87 & 76.03 & 78.37 \\
        PointFlow  &  75.68 & 70.74 & 60.65 & 62.36 & 62.84 & 60.57 \\
        SoftFlow  & 76.05 & 65.80 & 62.35 & 54.48 & 59.21 & 60.05 \\
        DPF-Net & 75.18 & 65.55 & 62.35 & 54.48 & 62.00 & 58.53 \\
        SetVAE & 76.54 & 67.65 & 59.95 & 59.94 & 58.84 & 60.57 \\
        SPAGHETTI & 78.20 & 77.00 & 72.30 & 71.00 & 70.70 & 69.00 \\
        DPM & 76.42 & 86.91 & 68.89 & 79.97 & 60.05 & 74.77 \\
        Shape-GF  & 80.00 & 76.17 & 63.20 & 56.53 & 68.96 & 65.48 \\
        CanonicalVAE & 80.15 & 76.27 & 63.23 & 61.56 & 62.78 & 61.05 \\
        PVD & 73.82 & 64.81 & 54.55 & 53.83 & 56.26 & 53.32 \\
        LION  & \underline{67.41} & 61.23 & \textbf{53.41} & \underline{51.14} & \textbf{53.70} & \underline{52.34} \\
        \midrule
        PointGPT  & 74.85 & 65.61 & 55.91 & 54.24 & 57.24 & 55.01 \\
        PSF  & 71.11 & \underline{61.09} & 57.19 & 56.07 & 58.92 & 54.45 \\
        SALAD  & 73.90 & 71.10 & 59.20 & 57.20 & 57.80 & 58.40 \\
        NSOT  & 68.64 & 61.85 & 59.66 & 53.55 & 55.51 & 57.63 \\
        DiPT  & -- & 74.32 & -- & 60.26 & -- & 64.47 \\
        TokenFlow (ours) & \textbf{66.65} & \textbf{60.55} & \underline{53.44} & \textbf{50.87} & \underline{54.26} & \textbf{52.21} \\
        \bottomrule
    \end{tabular}
    }
    \caption{Generation results on Airplane, Car, and Chair compared with baselines using 1-NNA as the metric. Lower is better. For clarity, the table is divided into two splits: the bottom corresponds to results from the past three years while the top corresponds to the earlier methods.}
    \label{tab:1-NN_generation_results}
\end{table*}

\paragraph{\textbf{Latent distribution and decoding.}}
A linear head turns each token into the parameters of a diagonal Gaussian.
\begin{align}
(\mu_m, \log \sigma_m) &= \mathrm{Head}(z_m), \\
q_\varphi(z_m|X) &= \mathcal{N}\!\left(z_m;\, \mu_m,\, \mathrm{diag}(\sigma_m^2)\right),
\end{align}
The latent tokens are sampled using the reparameterization trick \cite{yang2019pointflow,luo2021diffusion}. For decoding, we initialize $Y_0\sim\mathcal{N}(0,I)$ and integrate the conditional flow in Sec.~\ref{sec:rfm} conditioned on $Z$, yielding the reconstructed point cloud $\hat{X}=D_\psi(Z,Y_0)$. Since the encoder consists only of point-wise operations and permutation-invariant aggregation, $Q_\varphi(Z|X)$ is invariant to point ordering, while the decoder remains permutation equivariant.

\subsection{Prior Rectified Flow Matching}
\label{sec:rfm_prior}
Both levels of Eq.~\eqref{eq:ELBO} are trained by the regression of Sec.~\ref{sec:cfm} rather than by CNF likelihoods. The first level is a rectified flow over the latent token set, which supplies the shape-level representation that the second level is conditioned on.

\paragraph{\textbf{Flow over the latent set.}}
Take $Z_1 \sim q_\varphi(\cdot|X)$ from the encoder posterior and $Z_0 \sim \mathcal{N}(0, I)$. Along $Z_t = (1-t) Z_0 + t Z_1$ we regress
\begin{align}
\mathcal{L}_{\mathrm{prior}} = \mathbb{E}_{t, Z_0, Z_1} \big\| v_\theta(Z_t, t) - (Z_1 - Z_0) \big\|^2 ,
\label{eq:prior_loss}
\end{align}
where $v_\theta$ is a permutation-equivariant transformer over the tokens, conditioned by adaptive layer normalization on $t$.

\paragraph{\textbf{Optimal-transport coupling.}}
Trajectory straightness, and hence the number of Euler steps needed at inference, depends on how $Z_0$ is paired with $Z_1$. Following \cite{tong2023improving}, we compute an optimal-transport plan $\pi^\star$ between noise and data tokens within each minibatch, minimizing $\sum_{ij} \pi_{ij} \|Z_0^{(i)} - Z_1^{(j)}\|^2$, and draw training pairs from it. This lowers the variance of the regression target and straightens the flows.
\subsection{Conditional Rectified Flow Matching}
\label{sec:rfm}
The second level generates the coordinates themselves, conditioned on the latent set. They come from a conditional rectified flow starting at $Y_0 \sim \mathcal{N}(0, I)$:
\begin{align}
Y_t = (1-t) Y_0 + t X, \qquad
\frac{d Y_t}{d t} = v_\psi(Y_t, t; Z),
\label{eq:point_flow}
\end{align}
trained by
\begin{align}
\mathcal{L}_{\mathrm{cond}} = \mathbb{E}_{t, Y_0, X, Z} \big\| v_\psi(Y_t, t; Z) - (X - Y_0) \big\|^2 .
\label{eq:cond_loss}
\end{align}
It remains to specify $v_\psi$, which is a per-point network conditioned on the latent set. A shared MLP embeds each point, a cross-attention layer lets it read part-level information from the tokens $Z$, and the pooled shape code $\bar{z} = \mathrm{Pool}(Z)$ enters every point through adaptive layer normalization together with $t$:
\begin{align}
v_\psi(Y_t, t; Z)_i &= W_g \; \mathrm{AdaLN}\!\left( \mathrm{MLP}(y_{t,i});\, \bar{z},\, t \right) \notag \\
&+ W_l \; \mathrm{CrossAttn}\!\left( \mathrm{MLP}(y_{t,i}),\, Z,\, Z \right),
\label{eq:vpsi}
\end{align}
where
\begin{align}
\mathrm{AdaLN}(h; \bar{z}, t) = \gamma(\bar{z}, t) \odot \frac{h - \mu(h)}{\sigma(h)} \;+\; \beta(\bar{z}, t),
\label{eq:adaln}
\end{align}
with $(\gamma, \beta) = \mathrm{Affine}\big([\mathrm{LN}(\bar{z});\, \mathrm{emb}(t)]\big)$, layer normalization \cite{ba2016layer} and a sinusoidal time embedding, as is standard in diffusion transformers \cite{mo2023dit,zhang20233dshape2vecset}. The adaptive-normalization term carries the shape code into all points, so the field stays globally consistent, while the cross-attention term recovers the part-level detail that the pooled code does not retain.

\begin{figure*}[!tb]
    \centering
    \includegraphics[width=\textwidth]{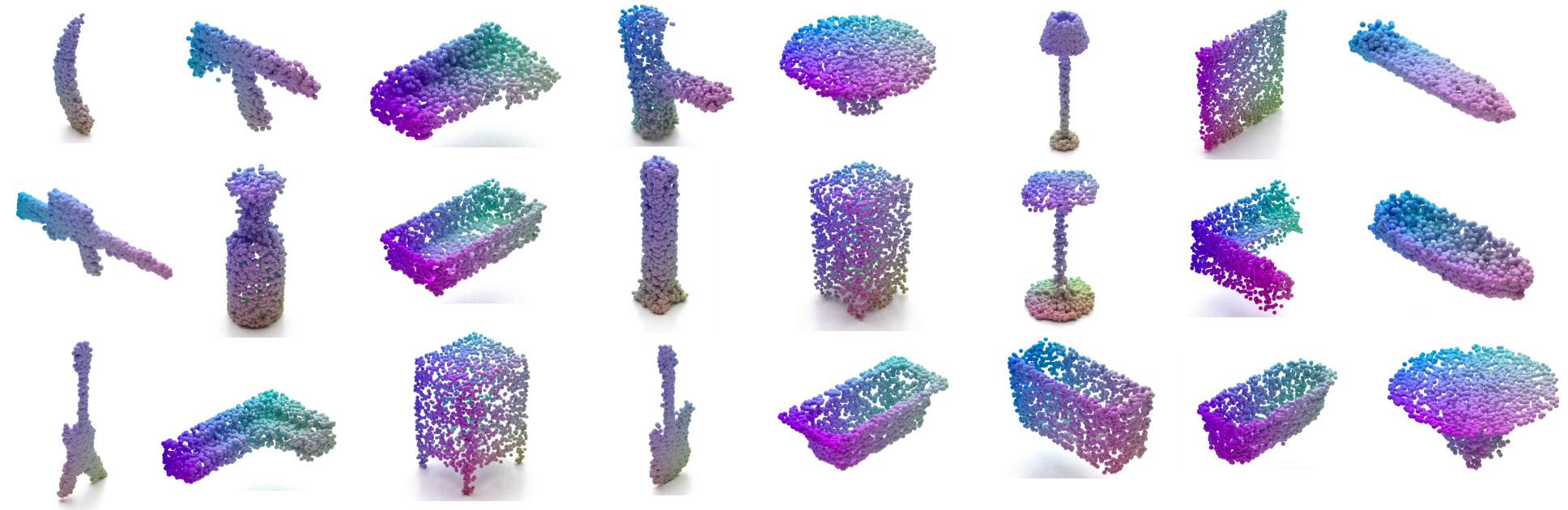}
    \caption{Multi-class results generated by TokenFlow.}
    \label{fig:multiclass_results}
\end{figure*}

\subsection{Objective Function}
\label{sec:obj}
All parameters are trained jointly. We keep the two-level ELBO of Eq.~\eqref{eq:ELBO} but swap PointFlow's  exact-likelihood terms \cite{yang2019pointflow} for the regressions of Eqs.~\eqref{eq:prior_loss} and \eqref{eq:cond_loss}, retaining its entropy term:
\begin{align}
\mathcal{L}_{\mathrm{total}} &= \lambda_p \mathcal{L}_{\mathrm{prior}}
+ \lambda_c \mathcal{L}_{\mathrm{cond}} \notag \\
&\quad + \lambda_{kl} \mathcal{L}_{KL}
- \lambda_{H} \mathcal{L}_{entro} ,
\label{eq:total}
\end{align}
where
\begin{align}
\mathcal{L}_{KL} &= \sum_{m=1}^{M} D_{KL}\!\big(\mathcal{N}(\mu_m, \sigma_m^2) \,\parallel\, \mathcal{N}(0, I)\big), \\
\mathcal{L}_{entro} &= H\big[Q_\varphi(Z|X)\big].
\end{align}
$\mathcal{L}_{\mathrm{cond}}$ keeps the geometry needed to rebuild the shape and anchors the code, $\mathcal{L}_{\mathrm{prior}}$ and $\mathcal{L}_{KL}$ align the aggregated posterior with a standard Gaussian so sampling is possible, and $\mathcal{L}_{entro}$ stops the posterior collapsing onto isolated codes, which would destroy diversity. Alg.~\ref{alg:train} lists the operations of one training step.

\begin{algorithm}[t]
\caption{One training step of TokenFlow.}
\label{alg:train}
\begin{algorithmic}[1]
\REQUIRE point cloud $X$, token count $M$, weights $\lambda_p, \lambda_c, \lambda_{kl}, \lambda_H$
\STATE $H \leftarrow \mathrm{ReLU}\big(\mathrm{Conv1d}(X)\big)$ \COMMENT{$(B, N, C)$}
\STATE $Z \leftarrow \mathrm{CBA}(H)$
\STATE $(\mu, \log \sigma) \leftarrow \mathrm{Head}(Z)$
\STATE $Z_1 \leftarrow \mu + \sigma \odot \epsilon$, \quad $\epsilon \sim \mathcal{N}(0, I)$
\STATE $Z_0 \sim \mathcal{N}(0, I)$
\STATE $\pi \leftarrow \arg\min_{\pi} \sum_{ij} \pi_{ij} \| Z_0^{(i)} - Z_1^{(j)} \|^2$
\STATE $Z_0 \leftarrow Z_0[\pi]$
\STATE $t_p \sim \mathcal{U}[0, 1]$, \quad $Z_t \leftarrow (1 - t_p) Z_0 + t_p Z_1$
\STATE $\mathcal{L}_{\mathrm{prior}} \leftarrow \| v_\theta(Z_t, t_p) - (Z_1 - Z_0) \|^2$
\STATE $Y_0 \sim \mathcal{N}(0, I)$, \quad $t_c \sim \mathcal{U}[0, 1]$, \quad $Y_t \leftarrow (1 - t_c) Y_0 + t_c X$
\STATE $\mathcal{L}_{\mathrm{cond}} \leftarrow \| v_\psi(Y_t, t_c; Z_1) - (X - Y_0) \|^2$ 
\STATE $\mathcal{L}_{KL} \leftarrow \sum_{m} D_{KL}\big(\mathcal{N}(\mu_m, \sigma_m^2) \,\|\, \mathcal{N}(0, I)\big)$
\STATE $\mathcal{L}_{entro} \leftarrow H\big[Q_\varphi(Z|X)\big]$
\STATE $\mathcal{L} \leftarrow \lambda_p \mathcal{L}_{\mathrm{prior}} + \lambda_c \mathcal{L}_{\mathrm{cond}} + \lambda_{kl} \mathcal{L}_{KL} - \lambda_H \mathcal{L}_{entro}$
\STATE Backpropagate $\mathcal{L}$
\end{algorithmic}
\end{algorithm}

\subsection{Complexity and Inference}
\label{sec:complexity}
\paragraph{Asymptotic cost.}
CBA cross-attention is the only one super-linear operation in the encoder at $O(NMC)$ time and $O(NM)$ memory, against $O(N^2 C)$ and $O(N^2)$ for Transformer self-attention (Tab.~\ref{tab:complexity}). Self-attention among the $M$ tokens adds $O(M^2 C)$, negligible when $M \ll N$. The prior flow acts on $M$ tokens and so is independent of $N$, and the conditional flow is point-wise in its input, so generation cost is linear in the number of points requested.
\paragraph{Inference.}
Sampling is a two-stage integration. The prior flow is integrated first, from $Z_0 \sim \mathcal{N}(0, I)$ to the latent set $Z_1$ using $K_p$ Euler steps of size $1/K_p$ at the left endpoints. The conditional flow is then integrated from an independently drawn $Y_0 \sim \mathcal{N}(0, I)$ conditioned on $Z_1$, using $K_c$ steps of the same scheme. Both step counts are set to 20, which is enough because the paths are trained to be straight. Alg.~\ref{alg:sample} summarizes this sampling procedure.

\begin{algorithm}[t]
\caption{Sampling.}
\label{alg:sample}
\begin{algorithmic}[1]
\REQUIRE Euler steps $K_p, K_c$, point count $N'$, trained fields $v_\theta, v_\psi$
\ENSURE point cloud $\hat{X} \in \mathbb{R}^{N' \times 3}$
\STATE $Z \sim \mathcal{N}(0, I)$ \COMMENT{$(M, d)$}
\FOR{$i = 0$ \TO $K_p - 1$}
\STATE $Z \leftarrow Z + \frac{1}{K_p} \, v_\theta\!\left(Z, \frac{i}{K_p}\right)$
\ENDFOR
\STATE $Y \sim \mathcal{N}(0, I)$ \COMMENT{$(N', 3)$}
\FOR{$i = 0$ \TO $K_c - 1$}
\STATE $Y \leftarrow Y + \frac{1}{K_c} \, v_\psi\!\left(Y, \frac{i}{K_c}; Z\right)$
\ENDFOR
\RETURN $Y$
\end{algorithmic}
\end{algorithm}
\section{Experiment}
The experiment section has four parts. Sec.~\ref{sec:evaluation_metrics} introduces the experimental setup, including baselines, evaluation metrics, configuration and implementation details. Sec.~\ref{sec:mainresults} reports the generation results against prior state-of-the-art methods. Sec.~\ref{sec:rle} evaluate the representation learning ability of the proposed autoencoders,
and Sec.~\ref{sec:ablation} ablates the proposed modules.
\subsection{Experimental Setup}
\label{sec:evaluation_metrics}
\paragraph{Baselines.} we compare with a variety of previous methods published on top conferences or journals. For generation task: r-GAN~\cite{achlioptas2018learning}, l-GAN (CD/EMD)~\cite{achlioptas2018learning}, PC-GAN~\cite{li2018point}, PointFlow~\cite{yang2019pointflow}, SoftFlow~\cite{kim2020softflow}, DPF-Net~\cite{klokov2020discrete}, SetVAE~\cite{kim2021setvae}, DPM~\cite{luo2021diffusion}, Shape-GF~\cite{cai2020learning}, CanonicalVAE~\cite{cheng2022autoregressive}, PVD~\cite{zhou20213d}, GET3D~\cite{gao2022get3d}, SPAGHETTI~\cite{hertz2022spaghetti}, MeshDiffusion~\cite{Liu2023MeshDiffusion}, PSF~\cite{wu2023fast}, LION~\cite{vahdat2022lion}, PointGPT~\cite{chen2023pointgpt}, SALAD~\cite{koo2023salad}, NSOT~\cite{hui2025not}, and DiPT~\cite{bastico2026rethinking}.
For representation learning task, we compare with: SPH~\cite{kazhdan2003rotation}, LFD~\cite{chen2003visual}, T-L Network~\cite{girdhar2016learning}, Vconv-DAE~\cite{sharma2016vconv}, PointGrow~\cite{sun2020pointgrow}, 3D-GAN~\cite{wu2016learning}, PointFlow~\cite{yang2019pointflow}, DPM~\cite{luo2021diffusion}, DualGAN~\cite{wen2021learning}, and TGN~\cite{xu2023transformer}.

\paragraph{Evaluation Metrics.} We employ COV (coverage), MMD (minimum matching distance), JSD (Jensen-Shannon divergence), and 1-NNA (1-nearest neighbor accuracy) using Chamfer Distance (CD) and Earth Mover's Distance (EMD) as base metrics, following prior works \cite{yang2019pointflow, klokov2020discrete}. COV detects mode collapse, MMD evaluates fidelity, JSD compares marginal distributions, and 1-NNA is a more robust measure for both diversity and quality~\cite{yang2019pointflow}.

\paragraph{Configuration.} For the generation task, we follow previous works \cite{yang2019pointflow,cai2020learning,kim2020softflow,klokov2020discrete,zhou20213d} and select three classes from ShapeNet dataset \cite{chang2015shapenet}: airplane, chair, and car. Each class in ShapeNet contains 15,000 sampled points. We sample 2,048 points for training and testing for each shape and preprocess the point cloud data following the steps outlined in \cite{yang2019pointflow}. 

For the classification evaluation of our auto-encoders, we implement ModelNet40 and ModelNet10 datasets \cite{wu20153d}, which contain 3D models of 40 and 10 categories, respectively.

\paragraph{Implementation Details.}
All experiments are conducted on 4 NVIDIA A800 80G GPUs. The point-wise embedding uses $C=128$ channels, and the CBA bottleneck uses $M=64$ learnable tokens of width $d=128$, $L_b=2$ blocks and four attention heads. The conditional point field has a hidden width of 128, and the latent velocity field is a four-layer Transformer. We optimize with Adam at a learning rate of $2\times10^{-3}$, with $\beta_1=0.9$, $\beta_2=0.999$ and no weight decay, for $4000$ epochs at a batch size of $128$. The learning rate stays constant over the first half of training and then decays linearly to zero. All four weights in Eq.~\eqref{eq:total} are set to $1$, and the minibatch optimal-transport coupling is enabled for the prior flow. Sampling uses $K_p=K_c=20$ Euler steps.

\subsection{Point Cloud Generation}
\label{sec:mainresults}
\paragraph{\textbf{Qualitative comparisons}} Fig.~\ref{fig:gen} compares samples from our model with those of prior methods.
We also train our model on all 55 different categories from the ShapeNet dataset. Fig.~\ref{fig:multiclass_results} shows multi-class point clouds generated by TokenFlow to demonstrate its generalizability.
\paragraph{\textbf{Quantitative comparisons}} We compare against prior state-of-the-art methods using the metrics of Sec.~\ref{sec:evaluation_metrics}. Results are given in Tab.~\ref{tab:other_metric_results} (MMD, COV, JSD) and Tab.~\ref{tab:1-NN_generation_results} (1-NNA). 
On the distributional metrics (Tab.~\ref{tab:other_metric_results}), TokenFlow attains almost the best results across all three categories. 

On 1-NNA (Tab.~\ref{tab:1-NN_generation_results}) TokenFlow achieves the best on Aiplane-CD/EMD, Car-EMD and Chair-EMD, while Car-CD and Chair-CD results are comparable with LION~\cite{vahdat2022lion}, which is a strong baseline. 

We note that fewer comparisons are reported in Tab.~\ref{tab:other_metric_results}, as several methods do not provide official results for MMD, COV or JSD. In contrast, 1-NNA has recently become a more widely adopted metric as shown in Tab. \ref{tab:1-NN_generation_results}.

\subsection{Unsupervised Representation Learning}
\label{sec:rle}
To additionally evaluate the representation learning ability of auto-encoders, we conduct classification experiments following previous methods \cite{yang2019pointflow,xu2023transformer,luo2021diffusion}. 
We first train TokenFlow with the full ShapeNet dataset and then train a linear SVM with extracted latent representations for classification. 
The classification results are summarized in Tab.~\ref{tab:modelnet_cls}. 
Our method achieves the best performance on ModelNet10 dataset, while demonstrating competitive results on the ModelNet40 dataset.
Further, we visualize the interpolation and extrapolation in the latent codes shown in Fig.~\ref{fig:inter_extra}.

\subsection{Ablation Study}
\label{sec:ablation}
We ablate the two core designs, the context module of the encoder and the number of latent tokens $M$.
\begin{table}[t]
\centering
\scriptsize
\begin{tabular}{lcc}
\toprule
Context module & 1-NNA-CD $\downarrow$ & 1-NNA-EMD $\downarrow$ \\
\midrule
Transformer \cite{vaswani2017attention} & 72.43 & 68.91 \\
PointTransformer \cite{zhao2021point} & \underline{71.54} & \underline{67.96}\\
Performer \cite{choromanski2020rethinking} &76.23 & 75.60 \\
CBA (ours) & \textbf{66.65} & \textbf{60.55} \\
\bottomrule
\end{tabular}
\caption{Context module ablation on the Airplane category, reported as 1-NNA (\%). Lower is better. Only the context module of the encoder is replaced, and everything else is held fixed.}
\label{tab:ablation}
\end{table}

\noindent \textbf{Context Module.} Tab.~\ref{tab:ablation} replaces the context module while the rest of the encoder and both flows stay unchanged. CBA improves on all three alternatives. PointTransformer is the closest competitor and still trails by 4.89 points on 1-NNA-CD and 7.41 points on 1-NNA-EMD. Transformer self-attention trails by 5.78 and 8.36 points, and Performer trails by 9.58 and 15.05 points.

The ranking follows from the way each module turns $N$ points into a shape code. Transformer self-attention mixes all point pairs and then has to pool the resulting $N$ features into one vector, which discards the assignment of points to parts. The conditional flow then has to recover that structure from noise, which is the hardest part of the problem. CBA keeps the assignment because its softmax is taken with $M$ learned queries over all $N$ points, so each token is a content-adaptive summary rather than a fixed average. 

PointTransformer does better than full attention because it retains local geometry, but its neighbourhood graph is built on Euclidean distance in the input coordinates. Parts are frequently not proximal, so the four legs of a chair are never placed in one group. 

Performer fails for a different reason. The kernel replaces the exponential, so attention mass spreads over many points and no cluster is selected. Its $M$ tokens then summarise diffuse second-order statistics instead of distinct parts.

The costs of the four modules are listed in Tab.~\ref{tab:complexity}. CBA replaces an $N \times N$ attention map with an $M \times N$ one, which is 32 times smaller at $N = 2048$ and $M = 64$. Performer builds no such map and is therefore the only module with a smaller memory footprint, but it is also the least accurate here, so the saving is paid for in fidelity rather than obtained for free.

\begin{table}[t]
\centering
\scriptsize
\begin{tabular}{lcc}
\toprule
Latent Tokens & 1-NNA-CD $\downarrow$ & 1-NNA-EMD $\downarrow$ \\
\midrule
$M = 1$ & 81.59 & 80.72 \\
$M = 8$ & 76.57 & 73.68\\
$M = 32$ & 72.30 & 67.84\\
$M = 64$ & \textbf{66.65} & \textbf{60.55} \\
$M = 128$ & \underline{70.03}& \underline{66.49} \\ 
\bottomrule
\end{tabular}
\caption{Effect of the number of latent tokens $M$ on the Airplane category, reported as 1-NNA (\%). Lower is better. $M = 64$ is the default used in the main results.}
\label{tab:ablation_m}
\end{table}

\noindent \textbf{Latent Tokens.} Tab.~\ref{tab:ablation_m} varies $M$. Accuracy improves from $M = 1$ to $M = 64$ and then degrades again at $M = 128$. At $M = 1$ the token self-attention becomes an identity, so the encoder degenerates into the single vector code used by earlier two-level models.

Small values of $M$ in general under-resolve the shape, because too few tokens cannot cover all parts. At $M = 128$ the opposite failure appears.
Tokens beyond the number of parts are redundant, and the latent prior must model a higher dimensional token set from the same training data, so the aggregated posterior drifts further from the Gaussian prior and sampling quality drops. The KL term is summed over every token coordinate, so its magnitude grows with $M$ as well. $M = 64$ balances the two effects and is the value used in the main results.
\section{Conclusion}
\label{sec:conclusion}

We proposed TokenFlow, a hierarchical rectified flow matching framework that
represents a shape as a set of $M \ll N$ latent tokens and couples a rectified
flow over these tokens with a conditional flow over points. We found that
existing models compress a shape into a single global vector, leaving part-level
structure implicit. To address
this, we introduced a context bottleneck that aggregates global
information at $O(NM)$ cost and conditions a point-level rectified flow. Experiments on ShapeNet demonstrate that our method achieves superior fidelity and diversity, with competitive transferable latent representations for ModelNet classification. 
Future work will extend the framework to large-scale scene-level point clouds and multimodal conditional generation.\\

{
    \small
    \bibliographystyle{ieeenat_fullname}
    \bibliography{main}
}


\end{document}